\documentclass[11pt]{article}
\usepackage{amsmath}
\usepackage{amssymb}
\usepackage{tcolorbox}
\usepackage[final]{acl}
\usepackage{algorithm}
\usepackage{algorithmic}
\usepackage{tabularx}
\usepackage{times}
\usepackage{latexsym}
\usepackage{booktabs}     
\usepackage{multirow}     
\usepackage{xcolor}

\usepackage{enumitem}    
\usepackage[T1]{fontenc}
\usepackage[utf8]{inputenc}
\usepackage{microtype}

\usepackage{inconsolata}

\usepackage{graphicx}

\title{Global Context or Local Detail? Adaptive Visual Grounding for Hallucination Mitigation}

\author{
Yubo Jiang$^{1,2}$ \quad
Xin Yang$^{2}$ \quad
Abudukelimu Wuerkaixi$^{2}$ \quad
Zheming Yuan$^{1}$ \quad
Xuxin Cheng$^{2}$ \\
\textbf{Fengying Xie}$^{1,3}$ \quad
\textbf{Zhiguo Jiang}$^{3}$ \quad
\textbf{Cao Liu}$^{2}$ \quad
\textbf{Ke Zeng}$^{2~\dagger}$ \quad
\textbf{Haopeng Zhang}$^{1,3~\dagger}$\\[0.5em]
$^{1}$School of Astronautics, Beihang University, Beijing 102206, China\\
$^{2}$Longcat Interaction Team, Meituan, Beijing 100102, China \\
$^{3}$Tianmushan Laboratory, Beihang University, Hangzhou 311115, China\\[0.5em]
{\tt\small \{jbond0409, zhanghaopeng\}@buaa.edu.cn (Y.J., H.Z.)}
}
\begin{document}
\maketitle
\begin{figure*}[t]
    \centering
    \includegraphics[width=0.96\textwidth]{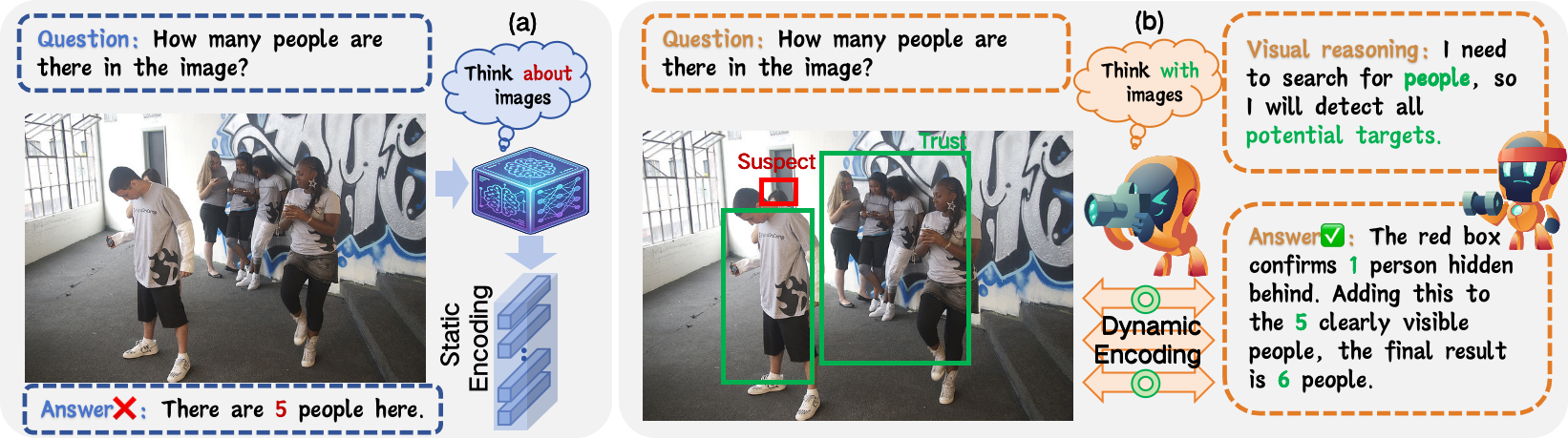}
    \caption{(a) Think about images: Traditional LVLMs rely on a one-time static encoding of the global image. (b) Think with images (Active-Look): Our TwI paradigm transforms the image into a real-time observation object. It actively generates auxiliary visual inputs through image manipulations to facilitate the model’s reasoning process.}
    \label{fig:motivation}
\end{figure*}
\begin{abstract}

Large vision--language models (LVLMs) excel at multimodal reasoning but still suffer from object-existence hallucinations when multi-step deliberation decouples from visual evidence. Think-with-Images (TwI) attempts to counter this by generating auxiliary observations (e.g., zoomed crops or highlighted views), yet it is not reliably beneficial. We identify two coupled failure modes: (1) a granularity--context trade-off of common operators, where zoom-in improves local detail but breaks global relations, while highlighting preserves topology but lacks fine evidence; and (2) an over-trust issue in tool-guided region proposals, where mislocalized evidence can dominate reasoning and even underperform standard prompting.
We propose Active-Look, a training-free, plug-and-play TwI framework that allocates visual computation by uncertainty. Active-Look runs two heterogeneous grounding experts in parallel and uses their disagreement as a proxy for uncertainty, spending the budget only to verify disputed regions. It further mitigates the operator trade-off with conflict-aware hybrid rendering: highlighting retains global context, while selective zoom-in performs local verification. Experiments on hallucination-focused and general benchmarks (POPE, MME, and CHAIR) across multiple LVLM backbones show consistent gains. 
\end{abstract}

\begingroup
\stepcounter{footnote}\footnotetext{$\dagger$ indicates the corresponding authors.}
\endgroup
\section{Introduction}

Large Vision--Language Models (LVLMs) have recently advanced multimodal reasoning by extending chain-of-thought style deliberation to visual inputs \cite{zhang2024vision,liu2025survey,wei2022chain}. 
Yet a persistent failure mode remains: textual reasoning can drift away from visual evidence, causing object-existence hallucinations---models confidently claim objects or attributes that are absent from the image \cite{lyu2023faithful,leng2024mitigating,rohrbach2018object}. 
One key reason is the one-shot perception pipeline: the image is encoded once into a static representation, and subsequent multi-step reasoning proceeds mostly in the language space with limited opportunities to re-check uncertain details (Figure~\ref{fig:motivation}a).

Recent work advocates Think-with-Images (TwI), where the model actively generates auxiliary observations (e.g., zoomed crops or highlighted regions) to support reasoning \cite{su2025thinking,zhang2025survey,zhang2025thyme}. However, two obstacles limit TwI reliability. First, TwI operations face a structural granularity--context trade-off: zooming improves local evidence but can break global topology and relations needed for compositional reasoning, while global highlighting preserves context yet often lacks resolution for small or ambiguous objects. This tension leads to bifurcated behavior---the same operation may help one subset of samples but harm another \cite{su2025thinking,zhao2025pyvision}. Second, many TwI pipelines depend on a single perception tool to decide where to look; a tool mistake becomes a single point of failure. A naive fix---stacking multiple tools and taking their union---can inject noisy proposals; when tools disagree, false positives may accumulate, degrading evidence quality rather than improving it \cite{zhang2025landscape}. Moreover, although TwI is used to improve multimodal reasoning, existing work provides limited and fragmented evaluation on object-existence hallucinations, and a mechanistic understanding of when TwI helps or hurts remains underdeveloped \cite{liu2025survey,leng2024mitigating}.

Motivated by these observations, we propose \textbf{Active-Look}, a conflict-driven active verification framework for TwI that allocates visual computation based on uncertainty \cite{han2024llm,talebirad2023multi}.
Active-Look follows three principles:
(i) uncertainty from disagreement---run heterogeneous dual experts in parallel and use their disagreement to localize ambiguous regions;
(ii) selective verification under budget---trigger additional perception only for flagged regions to control visual tokens;
(iii) context-preserving evidence---combine local verification with global context retention to resolve fine details without losing relational structure (Figure~\ref{fig:motivation}b).

We first formalize TwI as budgeted visual evidence acquisition (Section~\ref{sec:method}) and conduct diagnostics that expose (i) the granularity--context trade-off between zoom and highlight, and (ii) TwI fragility under tool errors (Section~\ref{sec:error}).
We then introduce Active-Look and evaluate it on hallucination-focused and general multimodal benchmarks across multiple LVLM backbones, showing reductions in hallucination while maintaining or improving VQA performance \cite{he2025llm,li2024survey}.
Beyond gains, we provide mechanistic analyses explaining when naive multi-tool union fails and how conflict-aware arbitration stabilizes TwI. Our contributions are as follows:

\begin{itemize}[itemsep=2pt,topsep=0pt,parsep=0pt]
    \item We characterize a consistent granularity--context trade-off in TwI operations (zoom vs.\ highlight) and show it induces performance bifurcation across sample regimes.
    \item We propose Active-Look, a conflict-driven verification framework that uses dual-expert disagreement to trigger selective re-perception while preserving global context.
    \item We demonstrate consistent improvements on hallucination and general multimodal benchmarks, and provide mechanistic evidence for why naive multi-tool union can degrade performance and how conflict-aware arbitration mitigates it.
\end{itemize}

\section{Related Work}

\subsection{Large Vision--Language Models (LVLMs)}
Large vision--language models such as LLaVA \cite{liu2023visual}, Qwen3-VL \cite{bai2025qwen3vltechnicalreport}, and InternVL2 \cite{chen2024expanding} have pushed multimodal reasoning by scaling alignment and visual capacity. Despite differences in architecture and training recipes, most LVLMs still follow a single-pass perception pipeline: the image is encoded once and provided as a static context for subsequent decoding. This design makes it difficult to revisit uncertain visual evidence during multi-step reasoning, which can leave intermediate conclusions weakly grounded. Our work complements LVLM backbones by enabling selective evidence re-acquisition at inference time, without modifying model weights \cite{wu2023multimodal,zhou2022learning}.

\subsection{Object Hallucination Mitigation}
Object hallucination remains a critical challenge for LVLMs. Within inference-time mitigations, decoding-based methods like VCD \cite{leng2024mitigating} and OPERA \cite{huang2024opera} adjust token generation but neglect underlying visual ambiguities. Alternatively, tool-augmented frameworks such as Woodpecker \cite{yin2024woodpecker} offer effectiveness but incur significant overhead due to iterative interactions. In contrast, Active-Look targets the evidence level via a lightweight verification mechanism, selectively triggering additional perception only upon expert disagreement to ensure computational efficiency.

\subsection{Think-with-Images Paradigm}
Think-with-Images reframes inference as interactive evidence acquisition, where the model generates auxiliary visual inputs to support reasoning. OpenAI o3 \cite{openai2025competitiveprogramminglargereasoning} also demonstrates the promise of thinking with images, although its internal mechanism remains unclear \cite{su2025thinking,liu2025survey}. Common instantiations include zoom-in cropping to amplify small or subtle evidence and visual prompting or highlighting to preserve global context while steering attention. However, prior work often commits to a fixed operator and lacks controlled analysis of the granularity--context trade-off that emerges across different sample regimes. Tool-guided variants rely on a single detector, inheriting a brittle failure mode when the tool is wrong. Our work provides a unified TwI formalization and diagnostic analyses, and introduces a conflict-aware arbitration mechanism based on dual-expert disagreement to stabilize tool-guided visual thinking in a training-free, plug-and-play manner \cite{wei2022chain,zhang2024vision}.

\begin{figure*}[t]
    \centering
    \includegraphics[width=0.90\textwidth]{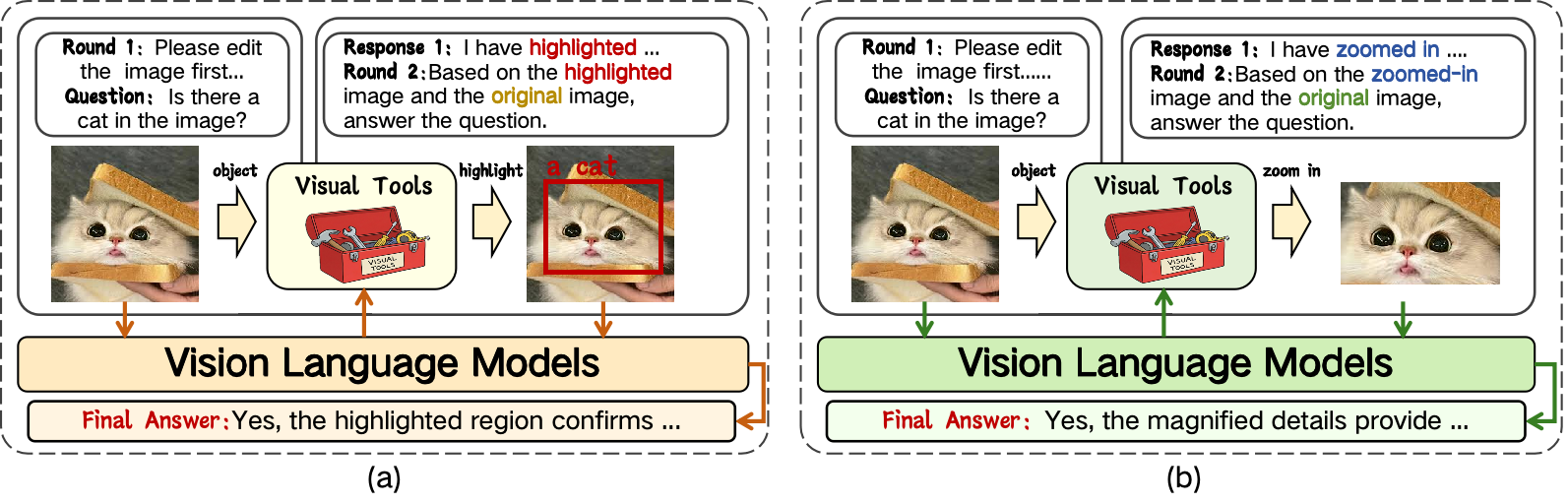}
    \caption{Two observation operators used in the Think-with-Images paradigm. 
(a) \textit{Highlight} preserves the full image and provides spatial cues to guide attention while keeping global context. 
(b) \textit{Zoom-in} crops a region of interest and enlarges it to increase visual detail, at the cost of removing surrounding context.}
    \label{fig:twi}
\end{figure*}

\section{Thinking with Images as Active Evidence Acquisition}
\label{sec:twi_prelim}

We present a unified view of \textit{Thinking with Images} (TwI) as a budgeted process of active visual evidence acquisition \cite{su2025thinking}. 
This section serves as preliminaries: it defines the iterative propose--select--render interface and two representative visual operators (Figure~\ref{fig:twi}). 
Our goal here is not to prescribe a particular selection algorithm, but to provide a common formulation for analyzing when different TwI instantiations help or hurt in later sections \cite{wei2022chain}.

\subsection{TwI as Budgeted Evidence Acquisition}
Given an image $I$ and a query $Q$, TwI constructs a sequence of auxiliary views $\{T_i\}_{i=1}^{N}$ by iteratively selecting regions of interest \cite{chen2023shikra}. 
At step $i$, the model maintains a textual reasoning state $R_{<i}$ and a set of previously observed visual embeddings $X_{<i}=\{X_1,\dots,X_{i-1}\}$.
A generic TwI step is written as:
\begin{equation}
\label{eq:twi_process}
\begin{split}
\mathcal{B}_i &= \operatorname{Propose}(I, Q, R_{<i}),\\
\mathbf{b}_i &= \operatorname{Select}(\mathcal{B}_i, Q, R_{<i}, X_{<i}),\\
T_i &= \Psi(I, \mathbf{b}_i),\quad X_i=\phi(T_i),\\
R_i &= \operatorname{LM}([X_i;\, R_{<i};\, Q]),
\end{split}
\end{equation}
where $\mathcal{B}_i$ is a candidate region set, $\mathbf{b}_i$ is the selected region, $\Psi$ is a visual operator that renders an auxiliary view, and $\phi(\cdot)$ is the visual encoder. 
TwI proceeds under a budget, e.g., a maximum number of steps $N$ or a cumulative token constraint:
\begin{equation}
\label{eq:twi_budget}
\sum_{i=1}^{N} c(\mathbf{b}_i) \le B,
\end{equation}
where $c(\mathbf{b}_i)$ denotes the visual token cost of encoding view $\Psi(I, \mathbf{b}_i)$. This constraint prioritizes efficient VLM context usage over exhaustive sampling, explicitly distinguishing variable reasoning costs from the fixed latency of tool execution.

\subsection{Region Proposal}
To avoid exhaustive scanning, TwI typically constructs $\mathcal{B}_i$ using lightweight region proposal \cite{liu2024grounding}. 
One instantiation uses a grounding model $\mathcal{G}$ conditioned on the current context:
\begin{equation}
\label{eq:twi_propose}
\mathcal{B}_i=\left\{\mathbf{b}\ \middle|\ \operatorname{score}_{\mathcal{G}}\!\left(\mathbf{b},\, \textsc{Query}(Q, R_{<i})\right)>\tau \right\}.
\end{equation}
This makes candidate regions aligned with the evolving hypothesis in $R_{<i}$, rather than arbitrary spatial crops.

\subsection{Selection Objective as a Conceptual Goal}
The role of $\operatorname{Select}(\cdot)$ is to choose regions that are most informative for answering $Q$ under the budget. 
This can be viewed as maximizing answer-relevant information gain while penalizing cost:
\begin{equation}
\label{eq:twi_select_concept}
\mathbf{b}_i=\arg\max_{\mathbf{b}\in\mathcal{B}_i}\ \Delta_i(\mathbf{b})-\lambda\, c(\mathbf{b}),
\end{equation}
where $\Delta_i(\mathbf{b})$ denotes the expected reduction of uncertainty about the answer after observing $\Psi(I,\mathbf{b})$. 
We treat Eq.~\eqref{eq:twi_select_concept} as an idealized objective: directly estimating information gain for black-box LVLMs is generally intractable, and practical TwI systems rely on computable proxies. 
This viewpoint enables a clean comparison across TwI instantiations, and will be used to motivate our later diagnostics.

\subsection{Visual Operators: Highlight and Zoom-in}
TwI differs mainly in how it renders the auxiliary view $T_i=\Psi(I,\mathbf{b}_i)$.
We focus on two widely used operators (Figure~\ref{fig:twi}), which expose a granularity--context trade-off \cite{peng2023kosmos}.

\paragraph{Highlight.}
Highlight preserves the global scene while providing spatial cues:
\begin{equation}
\label{eq:twi_high}
\Psi_{\text{high}}(I,\mathbf{b}_i)=\operatorname{Overlay}(I,\mathbf{b}_i).
\end{equation}
It retains global topology and relations, but may not improve perceptual detail for small objects.

\paragraph{Zoom-in.}
Zoom-in crops the region and resizes it to the model input resolution:
\begin{equation}
\label{eq:twi_zoom}
\Psi_{\text{zoom}}(I,\mathbf{b}_i)=\operatorname{Resize}(\operatorname{Crop}(I,\mathbf{b}_i)).
\end{equation}
Let $r(\mathbf{b})=\frac{|\mathbf{b}|}{HW}$ be the area ratio. Zooming increases effective resolution as $z(\mathbf{b})\approx r(\mathbf{b})^{-1/2}$, but may remove surrounding context that is important for compositional reasoning.

\subsection{Termination}
TwI terminates when the budget is exhausted, e.g., after $N$ steps or when $\sum_i c(\mathbf{b}_i)$ reaches $B$. 
The final answer is decoded from the accumulated multimodal trajectory $\{(X_i,R_i)\}_{i=1}^{N}$.

\section{Diagnosing TwI for Hallucination Mitigation}
\label{sec:error}

\begin{figure}[t]
    \centering
    \includegraphics[width=0.75\linewidth]{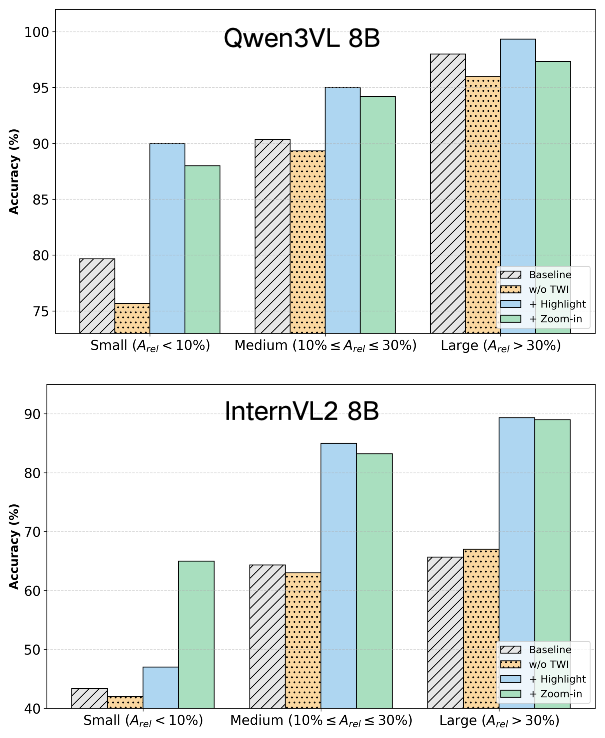} 
    \caption{\textbf{Scale-dependent effects of TwI operators.} Performance of Highlight, Zoom-in, and Prompting across object relative scales ($A_{rel}$) on Qwen3-VL-8B and InternVL2-8B.}
    \label{fig:scale_analysis}
\end{figure}

\begin{figure}[t]
    \centering
    \includegraphics[width=\linewidth]{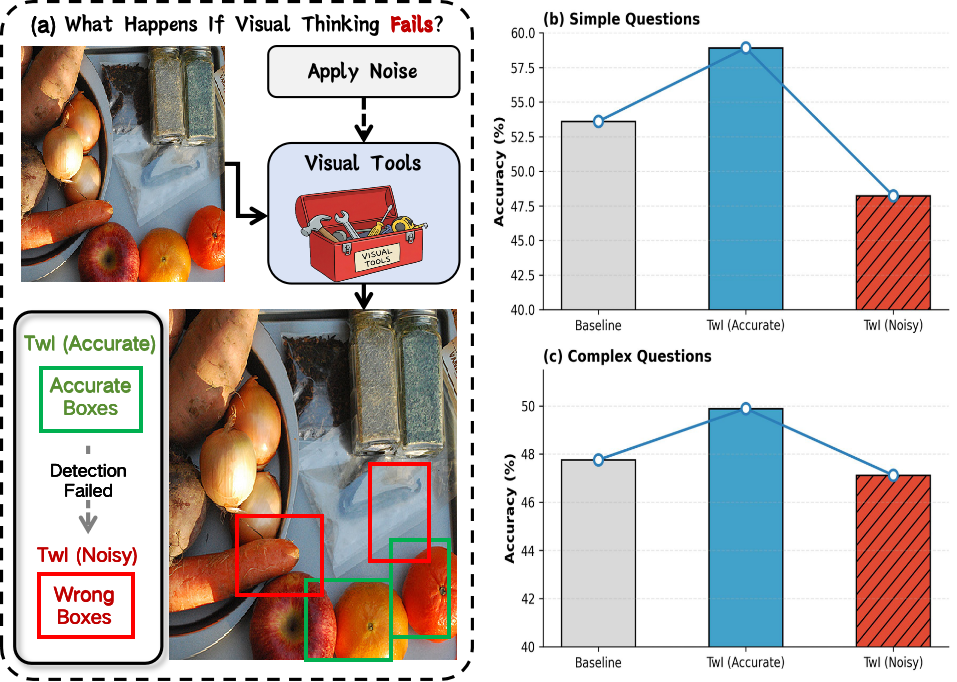} 
    \caption{\textbf{When visual guidance is unreliable.}
    (a) Spatial-noise injection to simulate proposal failures.
    (b--c) InternVL2-8B performance on TallyQA Simple and Complex subsets under accurate vs.\ noisy TwI guidance.}
    \label{fig:robustness_analysis}
\end{figure}

\paragraph{Diagnostics motivated by Section~\ref{sec:twi_prelim}.}
Section~\ref{sec:twi_prelim} frames Thinking-with-Images (TwI) as budgeted visual evidence acquisition, where each step selects an operator and a region proposal to render an auxiliary view. To isolate these effects under a controlled budget, we adopt a two-step setting ($N=2$) throughout this section: the model may perform at most one additional acquisition step before producing the final answer.

This formulation highlights two key factors:
(i) operator choice, reflecting a granularity--context trade-off, and
(ii) proposal reliability, determining whether rendered evidence is trustworthy.
Accordingly, we test two hypotheses:
H1: TwI gains vary across object scales due to the granularity--context trade-off;
H2: TwI becomes brittle under noisy proposals as models may over-trust externally rendered views.

\subsection{When Does TwI Help? A Scale-based Diagnosis}
\label{sec:scale_effect}
We examine whether TwI suppresses object-existence hallucinations
consistently across object scales. Following the POPE evaluation
format~\cite{li2023evaluating}, we adapt its question construction
and scoring to GQA~\cite{ainslie2023gqa}, enabling scale-controlled
diagnosis via bounding-box statistics. We evaluate Qwen3-VL-8B~\cite{bai2025qwen3vltechnicalreport}
and InternVL2-8B~\cite{chen2024expanding} under Highlight (spatial cueing),
Zoom-in (local re-rendering), and Prompting (reasoning-only control).
Target objects are grouped by relative area $A_{rel}$ into
Small ($<10\%$), Medium ($10\%$--$30\%$), and Large ($>30\%$)
following Supplementary Material~2. A TwI w/o Visual Feedback
ablation preserves the multi-step reasoning structure while
withholding newly rendered views.

\paragraph{Results.}
Figure~\ref{fig:scale_analysis} shows a clear scale-dependent
bifurcation. For InternVL2-8B, Zoom-in performs best on Small
objects while Highlight dominates on Large ones, reflecting the
granularity--context trade-off. Removing visual feedback causes
performance to fall below the static baseline, confirming that
TwI gains require additional visual evidence rather than
chain-of-thought prompting alone. For the stronger Qwen3-VL-8B,
Highlight remains consistently competitive and often outperforms
Zoom-in even on smaller objects, suggesting diminishing returns
from aggressive resampling when native visual encoding is strong.

\subsection{When Does TwI Hurt? Failure Modes under Noisy Visual Thoughts}
\label{sec:visual_failure_analysis}
Open-world settings inevitably involve imperfect region proposals.
We stress-test TwI under controlled proposal noise on
TallyQA~\cite{acharya2019tallyqa}, comparing Standard Prompting,
TwI (Accurate), and TwI (Noisy), where boxes are spatially shifted
to enforce low overlap ($IoU < 0.3$), across Simple and Complex subsets.

\paragraph{Results.}
As shown in Figure~\ref{fig:robustness_analysis}, TwI improves
performance under accurate guidance but degrades sharply under noise,
dropping below the static baseline in both subsets
(Simple: 48.2\% vs.\ 53.6\%; Complex: 47.1\% vs.\ 47.8\%),
revealing an over-trust failure mode: once rendered, erroneous
visual evidence can dominate subsequent reasoning.

\paragraph{Implications for a robust TwI framework.}
These diagnoses indicate that TwI is not inherently robust.
A fixed operator can misallocate visual budget across regimes,
and unverified proposals introduce a single point of failure.
These findings motivate Active-Look, a conflict-driven TwI framework
that verifies ambiguous visual evidence and adaptively balances
global context retention with local detail inspection under uncertainty.

\begin{figure*}[t]
    \centering
    \includegraphics[width=0.9\textwidth]{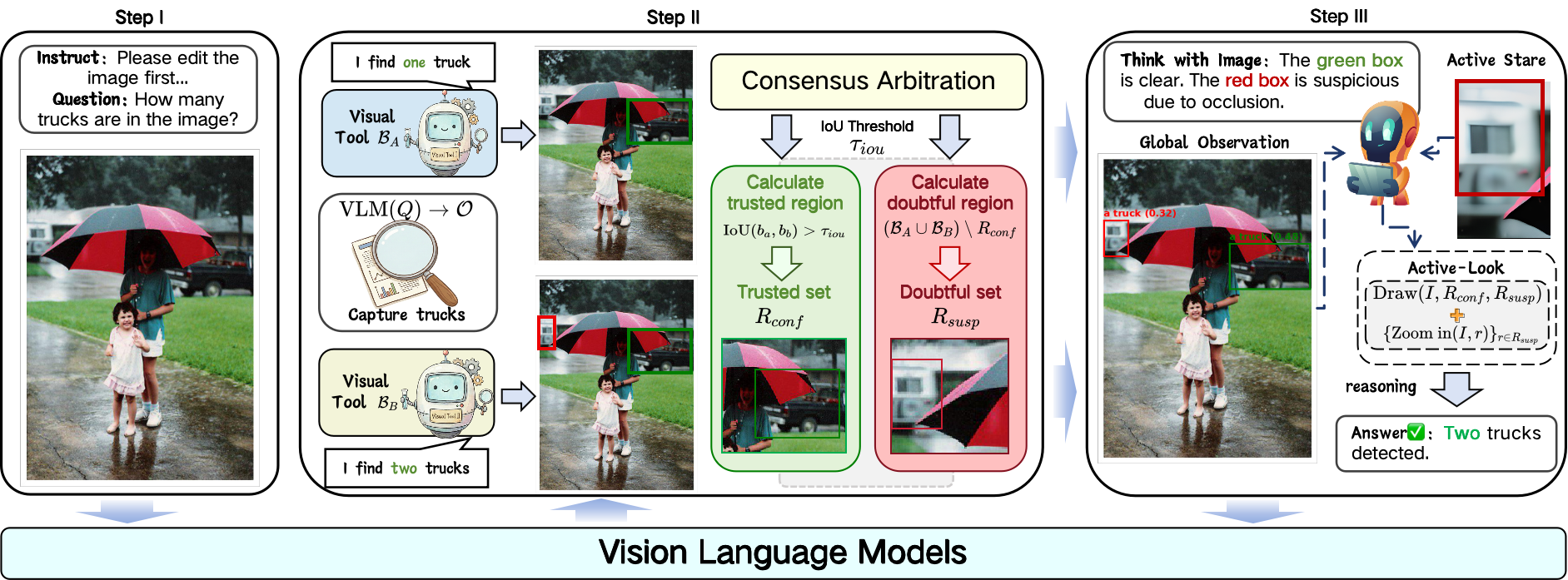}
    \caption{Overview of Active-Look. Given an image and a query, two heterogeneous visual tools propose candidate regions. A consensus arbitration module partitions them into a trusted set $R_{\mathrm{conf}}$ and a doubtful set $R_{\mathrm{susp}}$. Active-Look then renders a global highlighted view conditioned on $R_{\mathrm{conf}}$ and zoomed-in crops for $R_{\mathrm{susp}}$, allocating the budget to verify disputed evidence before decoding the final answer.}
    \label{fig:active_look}
    
\end{figure*}

\section{Methodology: Active-Look}
\label{sec:method}

Section~\ref{sec:error} reveals two coupled failure modes of existing Think-with-Images systems:
(1) a \emph{granularity--context} dilemma caused by a fixed operator choice $\Psi$ (Zoom-in may sever relations; Highlight may lack detail),  
(2) an \emph{over-trust} failure when region proposals are imperfect, where the LVLM follows mislocalized evidence and performs worse than static prompting.
In this section, we introduce Active-Look, a training-free TwI instantiation that makes the \emph{selection} step robust by prioritizing \emph{disputed} regions and allocating the budget to verification. Active-Look as a concrete instantiation of Eq.~\eqref{eq:twi_process}.
We follow the TwI interface in Eq.~\eqref{eq:twi_process}. Active-Look specifies:
(i) $\operatorname{Propose}(\cdot)$ via \emph{heterogeneous dual experts} to reduce single-detector bias;
(ii) $\operatorname{Select}(\cdot)$ via a computable proxy of answer uncertainty (expert disagreement), serving as a practical surrogate for the conceptual objective in Eq.~\eqref{eq:twi_select_concept};
(iii) $\Psi(\cdot)$ as a \emph{hybrid rendering} that combines $\Psi_{\text{high}}$ for global topology and $\Psi_{\text{zoom}}$ for local verification under a budget.

\begin{algorithm}[t]
\small
\caption{Active-Look (a training-free TwI instantiation)}
\label{alg:active_look}
\begin{algorithmic}
\REQUIRE Image $I$, Query $Q$, experts $\mathcal{G}_A,\mathcal{G}_B$, budget $B$.
\ENSURE Response $A$.
\vspace{2pt}

\STATE \textbf{1. Propose (Hypothesis-driven perception)}
\STATE Targets $\mathcal{O} \leftarrow \operatorname{LM}(Q)$.
\STATE $\mathcal{B}^A \leftarrow \operatorname{Propose}_{\mathcal{G}_A}(I,Q,\emptyset)$; \quad
      $\mathcal{B}^B \leftarrow \operatorname{Propose}_{\mathcal{G}_B}(I,Q,\emptyset)$.
\STATE $\mathcal{B} \leftarrow \mathcal{B}^A \cup \mathcal{B}^B$.

\STATE \textbf{2. Select (Consensus arbitration as a proxy)}
\STATE Compute conflict ratio $\gamma$ and adaptive threshold $\tau_{iou}$ (Eq.~\ref{eq:adaptive_threshold}).
\STATE Partition proposals into Trusted $R_{conf}$ and Doubtful $R_{susp}$ via IoU matching under $\tau_{iou}$.
\STATE Select zoom set $\mathcal{S}\subseteq R_{susp}$ by descending disagreement score, s.t.\ $\sum_{\mathbf{b}\in \mathcal{S}} c(\mathbf{b}) \le B$.

\STATE \textbf{3. Render (Conflict-aware active glance)}
\STATE $T_{gl} \leftarrow \Psi_{\text{high}}(I, R_{conf}\cup R_{susp})$. \hfill $\triangleright$ global highlight
\STATE $\mathcal{T}_{zm} \leftarrow \{ \Psi_{\text{zoom}}(I,\mathbf{b}) \}_{\mathbf{b}\in \mathcal{S}}$. \hfill $\triangleright$ selective zoom

\STATE \textbf{4. Reason (Multi-view inference)}
\STATE $X_{gl}\leftarrow \phi(T_{gl})$; \quad $\mathcal{X}_{zm}\leftarrow \{\phi(T)\mid T\in\mathcal{T}_{zm}\}$.
\RETURN $A \leftarrow \operatorname{LM}([\mathcal{X}_{zm};\, X_{gl};\, Q])$.
\end{algorithmic}
\end{algorithm}

\subsection{Hypothesis-Driven Perception}
Active-Look begins with hypothesis-driven perception, aligned with the TwI proposal step in Eq.~\eqref{eq:twi_process}.
Given $I$ and $Q$, we extract target concepts $\mathcal{O}$ and generate candidate regions with two heterogeneous grounding experts,
$\mathcal{G}_A$ (GroundingDINO\cite{liu2024grounding}) and $\mathcal{G}_B$ (OWLv2\cite{minderer2023scaling}).
Each expert instantiates Eq.~\eqref{eq:twi_propose} to produce proposal sets $\mathcal{B}^A$ and $\mathcal{B}^B$; we take their union $\mathcal{B}=\mathcal{B}^A\cup\mathcal{B}^B$ as the candidate pool,
reducing reliance on any single proposer.

\subsection{Simulating Human Verification: Glance vs. Stare}
Active-Look instantiates the \textit{Select} step with a computable uncertainty proxy: \emph{cross-expert disagreement} \cite{parr2017active,hino2023active}.
Regions proposed by only one expert tend to be more ambiguous, and thus more informative to verify---a practical surrogate for the conceptual objective in Eq.~\eqref{eq:twi_select_concept}.

\begin{table*}[t]
\centering
\renewcommand{\arraystretch}{0.70}
\begin{tabular*}{0.95\textwidth}{@{\extracolsep{\fill}}llcccccc@{}}
\toprule
\textbf{Model} & \textbf{Method} & \textbf{Accuracy} & \textbf{Precision} & \textbf{Recall} & \textbf{F1} \\
\midrule
\multirow{7}{*}{\textbf{LLaVA1.5-7B\cite{liu2023visual}}}
    & Baseline & 76.10 & 78.47 & 71.93 & 75.06  \\
    & VCD & 78.12 & 80.86 & 73.15 & 76.81  \\
    & OPERA & 79.89 & \textbf{85.12} & 73.04 & 78.62  \\
    & VAF & 78.96 & 81.88 & 74.95 & 78.26  \\
    & Highlight & 78.21 & 80.82 & 77.36 & 79.05  \\
    & Zoom-in & 78.43 & 81.13 & 78.32 & 79.70  \\
    & Active-Look & \textbf{80.55} & 82.41 & \textbf{79.29} & \textbf{80.82}  \\
\midrule
\multirow{4}{*}{\textbf{Qwen3-VL-8B\cite{bai2025qwen3vltechnicalreport}}}
    & Baseline & 84.32 & 84.78 & 83.37 & 84.07  \\
    & Highlight & 85.34 & 85.79 & 84.71 & 85.25 &  \\
    & Zoom-in & 84.42 & 84.92 & 84.12 & 84.52  \\
    & Active-Look & \textbf{89.26} & \textbf{89.09} & \textbf{87.85} & \textbf{88.46} \\
\midrule
\midrule
\multicolumn{6}{c}{\textit{More Model Results}} \\
\midrule
\midrule
\multirow{2}{*}{\textbf{LLaVA1.5-13B\cite{liu2023visual}}} 
    & Baseline & 77.33 & 81.68 & 70.47 & 75.66  \\
    & Active-Look & \textbf{81.25} & \textbf{85.93} & \textbf{74.86} & \textbf{80.01}  \\
\midrule
\multirow{2}{*}{\textbf{LLaVA-NeXt-7B\cite{li2024llava}}} 
    & Baseline & 81.85 & 88.19 & 72.64 & 79.66  \\
    & Active-Look & \textbf{85.73} & \textbf{92.81} & \textbf{77.47} & \textbf{84.45}  \\
\midrule
\multirow{2}{*}{\textbf{InternVL2-4B\cite{chen2024expanding}}} 
    & Baseline & 78.12 & 83.27 & 73.34 &  77.98  \\
    & Active-Look & \textbf{82.78} & \textbf{83.64} & \textbf{78.32} & \textbf{80.89} \\
\midrule
\multirow{2}{*}{\textbf{InternVL2-8B\cite{chen2024expanding}}} 
    & Baseline & 80.60 & \textbf{85.92} & 73.20 & 79.05  \\
    & Active-Look & \textbf{85.00} & 84.28 & \textbf{80.40} & \textbf{82.28} \\
\midrule
\multirow{2}{*}{\textbf{Qwen3-VL-4B\cite{bai2025qwen3vltechnicalreport}}} 
    & Baseline & 83.73 & 83.23 & 82.17 & 82.69  \\
    & Active-Look & \textbf{87.97} & \textbf{88.36} & \textbf{85.76} & \textbf{87.04}  \\

\bottomrule
\end{tabular*}
\caption{\textbf{Results on POPE.} We report Accuracy, Precision, Recall, and F1 under prompting baselines, TwI variants, and prior mitigation methods (VCD\cite{leng2024mitigating}
, OPERA\cite{huang2024opera}, VAF\cite{yin2025clearsight}).}
\label{tab:pope_results}
\end{table*}

\begin{table*}[t]
\centering
\renewcommand{\arraystretch}{0.7}
\setlength{\abovecaptionskip}{5pt}
\setlength{\belowcaptionskip}{5pt}
\begin{tabular*}{0.95\textwidth}{@{\extracolsep{\fill}}llcccccc@{}}
\toprule
\multirow{2}{*}{\textbf{Model}} & \multirow{2}{*}{\textbf{Method}} & \multicolumn{2}{c}{\textbf{Object-level}} & \multicolumn{2}{c}{\textbf{Attribute-level}} & \multirow{2}{*}{\textbf{Total Score}} \\
\cmidrule(lr){3-4} \cmidrule(lr){5-6}
 & & \textbf{Existence} & \textbf{Count} & \textbf{Position} & \textbf{Color} & \\
\midrule
\multirow{2}{*}{LLaVA-1.5-7B}
    & Baseline & 158.33 & 91.33 & 90.00 & 91.66 & 431.33 \\
    & VCD &  150.00& 98.33& 93.33& 104.33 &  446.00   \\
    & OPERA  & 168.33& 110.00& 113.66 & 105.33 & 497.33 \\
    & VAF  & 158.33 & 111.33 & 113.66 & 101.66 & 485.00 \\ 
    & Active-Look & 168.33 & \textbf{121.66} & \textbf{116.66} & \textbf{110.00} & \textbf{516.66} \\
\midrule
\multirow{2}{*}{InternVL2-8B}
    & Baseline & 180.00 & 128.33 & 120.00 & 140.00 & 568.33 \\
    & Active-Look & \textbf{185.00} & \textbf{130.00} & \textbf{131.66} & \textbf{180.00} & \textbf{626.66} \\
\midrule
\multirow{2}{*}{Qwen3-VL-8B}
    & Baseline & 190.00 & 165.00 & 143.33 & 175.00 & 673.33 \\
    & Active-Look & \textbf{195.00} & \textbf{170.00} & \textbf{148.33} & \textbf{185.00} & \textbf{698.33} \\
\bottomrule
\end{tabular*}
\caption{\textbf{Results on MME (Existence/Count/Position/Color).} We evaluate Active-Look on four hallucination-relevant categories and report per-category scores and the total score.}
\label{tab:MME}
\end{table*}

\paragraph{Consensus partition.}
We perform IoU-based matching between $\mathcal{B}^A$ and $\mathcal{B}^B$ and partition proposals into:
(i) \textbf{Trusted set} $R_{conf}$, and
(ii) \textbf{Doubtful set} $R_{susp}$.
Let $\gamma$ denote the scene conflict ratio 
$\gamma=\frac{|R_{susp}|}{|R_{conf}|+|R_{susp}|}$.
We then adapt the IoU threshold $\tau_{iou}$ based on $\gamma$:
\begin{equation}
\label{eq:adaptive_threshold}
    \tau_{iou} =
    \begin{cases}
    \tau_{base} - \delta & \text{if } \gamma < \tau_{low}   \\
    \tau_{base} + \delta & \text{if } \gamma > \tau_{high}  \\
    \tau_{base} & \text{otherwise.}
    \end{cases}
\end{equation}
This ``glance vs.\ stare'' logic makes selection budget-aware:
Trusted regions are handled via global context, while only doubtful regions consume additional budget for verification.

\paragraph{Budgeted selection within $R_{susp}$.}
Within $R_{susp}$, we prioritize regions with stronger disagreement signals and select a subset $\mathcal{S}\subseteq R_{susp}$ under the budget constraint $\sum_{\mathbf{b}\in\mathcal{S}} c(\mathbf{b})\le B$ (Eq.~\eqref{eq:twi_budget}).
This turns the intractable information-gain objective into a computable and robust proxy\cite{xu2023dynamic}.

\subsection{Conflict-Aware Active Glance}
Active-Look verifies disputed evidence via a hybrid rendering strategy that resolves the granularity--context trade-off (Section~\ref{sec:error}).

\paragraph{Hybrid rendering (Highlight + selective Zoom-in).}
We first render a global highlighted view
$T_{gl}=\Psi_{\text{high}}(I, R_{conf}\cup R_{susp})$ (Eq.~\eqref{eq:twi_high}),
preserving scene topology while cueing trusted and doubtful regions.
We then zoom in only on selected doubtful boxes $\mathbf{b}\in\mathcal{S}$:
$T_{\mathbf{b}}=\Psi_{\text{zoom}}(I,\mathbf{b})$ (Eq.~\eqref{eq:twi_zoom}),
focusing the budget on regions where extra granularity is most informative\cite{yuan2023small}.

\paragraph{Multi-view reasoning.}
We encode all views with $\phi(\cdot)$ (Eq.~\eqref{eq:twi_process}) and decode
$A\leftarrow \operatorname{LM}([\mathcal{X}_{zm}; X_{gl}; Q])$.
Joint conditioning on global context ($X_{gl}$) and verified local evidence ($\mathcal{X}_{zm}$) reduces hallucinations while remaining plug-and-play at inference time.

\section{Experiments}
\label{sec:experiments}
\subsection{Experimental Settings}
\label{subsec:settings}
\paragraph{Datasets.}
We evaluate on three representative benchmarks that cover hallucination diagnosis and general multimodal capability:

\textbf{POPE\cite{li2023evaluating}}targets object-existence hallucination via yes/no questions about whether specific objects appear in an image; we use the Adversarial split, constructed on MSCOCO\cite{lin2015microsoftcococommonobjects} images.

\textbf{MME\cite{fu2025mmecomprehensiveevaluationbenchmark}} is a comprehensive benchmark for multimodal perception and cognition; we report four hallucination-relevant categories: Existence, Count, Position, and Color.

\textbf{CHAIR\cite{objectHallucination}} measures object hallucination in image captioning by comparing objects mentioned in captions against ground-truth object annotations. We report CHAIR scores as the hallucination rate of mentioned objects.

\paragraph{VLM Backbones.}
We test Active-Look on three representative LVLMs: LLaVA\cite{liu2023visual}, Qwen3-VL\cite{bai2025qwen3vltechnicalreport}, and InternVL\cite{chen2024expanding}.
These backbones differ in vision encoders and vision--language fusion/alignment designs, enabling us to assess robustness across heterogeneous architectures.

\subsection{Experiments Results}

\paragraph{Results on POPE.} 
As shown in Table~\ref{tab:pope_results}, Active-Look consistently outperforms standard prompting and single-operator TwI variants (Highlight/Zoom-in) across all backbones. It yields notable gains on LLaVA-7B (+4.45\% Acc) and Qwen3-VL-8B (+4.94\% Acc), indicating that conflict-triggered verification mitigates hallucinations even in advanced LVLMs. On InternVL2-8B, Recall improves markedly ($73.20 \rightarrow 80.40$), suggesting that selective zooming can recover missed objects while preserving a strong precision--recall balance.

\paragraph{Results on MME.} 
Table~\ref{tab:MME} evaluates performance across four hallucination-prone categories: Existence, Count, Position, and Color. Active-Look yields consistent improvements in Total Scores, with particularly pronounced gains in fine-grained perception tasks. LLaVA-1.5-7B's Count and Position scores increase by 30.33 and 26.66, respectively. These results underscore the efficacy of our training-free perceptual enhancement, which strengthens detail-sensitive reasoning by refining local cues without sacrificing global context.

\begin{table}[t]
\centering
\setlength{\abovecaptionskip}{5pt}
\setlength{\belowcaptionskip}{5pt}
\setlength{\tabcolsep}{4pt}
\renewcommand{\arraystretch}{0.5}
\begin{tabular}{@{}llccc@{}}
\toprule
\textbf{Model} & \textbf{Methods} & \textbf{$\mathcal{C}$s} $\downarrow$ & \textbf{$\mathcal{C}$i} $\downarrow$ & \textbf{Recall} $\uparrow$ \\
\midrule
\multirow{2}{*}{LLaVA-1.5-7B}
    & Baseline & 53.0 & 18.4 & \textbf{67.0} \\
    & Active-Look & \textbf{15.0} & \textbf{15.8} & 64.9 \\
\addlinespace[2pt]
\multirow{2}{*}{InternVL2-8B}
    & Baseline & 37.0 & 9.3 & 62.3 \\
    & Active-Look & \textbf{21.5} & \textbf{6.6} & \textbf{64.7} \\
\addlinespace[2pt]
\multirow{2}{*}{Qwen3-VL-8B}
    & Baseline & 40.0 & 9.9 & \textbf{74.1} \\
    & Active-Look & \textbf{32.0} & \textbf{8.3} & 70.0 \\
\bottomrule
\end{tabular}
\caption{CHAIR evaluation on MSCOCO captioning.}
\label{tab:chair_comparison}
\end{table}

\paragraph{Results on CHAIR.}
To assess reliability in free-form generation, we evaluate Active-Look on CHAIRs (Table~\ref{tab:chair_comparison}). Active-Look consistently reduces object hallucinations, with the largest gain on LLaVA-1.5-7B, where sentence-level hallucination ($\mathcal{C}_S$) drops by 71.7\% relatively. Some models show a slight recall trade-off, whereas InternVL2-8B improves both $\mathcal{C}_S$ (37.0 $\rightarrow$ 21.5) and Recall (62.3 $\rightarrow$ 64.7). These results suggest that conflict-triggered verification suppresses non-existent object mentions and can also recover missed objects in stronger backbones.

\paragraph{Ablation Study}

\begin{table}[t]
\centering
\renewcommand{\arraystretch}{0.5}
\begin{tabular}{lcc|cc}
\toprule
\textbf{Tool A} & \textbf{Tool B} & \textbf{C\&C} & \textbf{Ac (\%)} & \textbf{F1 (\%)} \\ \midrule
\checkmark               &                          &                               & 87.12                & 86.69                 \\
                         & \checkmark               &                               & 87.87                & 87.28                 \\
\checkmark               & \checkmark               &                               & 86.14                & 85.84                 \\
\checkmark               & \checkmark               & \checkmark                    & \textbf{89.26}       & \textbf{88.46}        \\ \bottomrule
\end{tabular}
\caption{Ablation study of different visual experts and the Consensus + Conflict mechanism(C\&C). The baseline incorporates individual experts, while our full model (last row) achieves the best performance.}
\label{tab:ablation_study}
\end{table}

We analyze Table~\ref{tab:pope_results} and Table~\ref{tab:ablation_study} to validate the core design choices of Active-Look.

\paragraph{Synergistic Visual Cues: Highlight vs. Zooming.} 
Table~\ref{tab:pope_results} demonstrates that \textit{Active-Look} consistently outperforms single-operator TwI variants. While \textit{Highlighting} preserves global topology and \textit{Zoom-in} enhances local granularity, neither alone fully resolves the granularity-context trade-off. \textit{Active-Look} bridges this gap by adaptively triggering local zooms within a highlighted global frame, achieving a superior precision-recall balance that surpasses individual visual perspectives.

\paragraph{Efficacy of Conflict Arbitration.} 
Ablation studies in Table~\ref{tab:ablation_study} reveal that simply increasing the number of proposals can be counter-productive: a naive union of dual experts (\textit{Union A $\cup$ B}) yields lower accuracy (86.14\%) than a standalone Expert B (87.87\%). This counter-intuitive result suggests that unstructured aggregation propagates conflicting false positives, leading to over-confident yet erroneous decisions. \textit{Active-Look} mitigates this by treating expert disagreements as triggers for targeted verification, effectively transforming additional perceptual noise into resolved certainty.

\section{Conclusion}
In this work, we present an empirical study of Think-with-Images (TwI) and identify two key findings: (i) operator effectiveness (Highlight vs.\ Zoom-in) is regime-dependent, varying with object scale and reasoning difficulty, and (ii) TwI can become fragile under imperfect proposals due to an over-trust failure mode.
Motivated by these diagnostics, we propose Active-Look, a training-free framework that allocates perceptual resources based on uncertainty via a glance--stare mechanism.
Through Dual-Expert Consensus, Active-Look isolates visual-logical conflicts, preserves global topology with Highlight, and applies selective Zoom-in only where additional granularity is most informative.
Experiments across multiple benchmarks demonstrate improved hallucination mitigation and robustness under noisy guidance, suggesting conflict-driven evidence acquisition as a direction for reliable multimodal reasoning.

\section*{Acknowledgement}
This work is sponsored by Beijing Nova Program.

\section*{Limitations}
While Active-Look demonstrates significant effectiveness in mitigating object-existence hallucinations, we acknowledge three limitations. 
(I) Our framework relies on the recall capability of external grounding experts . Although the dual-expert strategy improves coverage, the system may still fail to verify targets that are missed by both detectors, particularly for extremely rare objects or abstract concepts.
(II) We clarify that the "budget" constraint $B$ defined in Eq.~(2) specifically targets the visual token consumption of the LVLM, rather than total system latency. 
While Active-Look optimizes the reasoning process by selectively zooming only into conflicted regions—thereby saving VLM context length compared to exhaustive cropping—the hypothesis-driven perception stage incurs a fixed computational overhead due to the execution of dual external experts. 
Therefore, our framework represents a strategic trade-off: we invest strictly budgeted pre-processing computation to ensure token economy and higher faithfulness, preventing the LVLM's context window from being overwhelmed by irrelevant visual information.
(III) Our current conflict-aware arbitration is optimized primarily for object existence and attribute verification. Extending this paradigm to correct hallucinations in complex spatial reasoning or action understanding remains a direction for future work.

\section*{Ethical Considerations}

We propose Active-Look to improve the faithfulness of LVLMs, yet we acknowledge specific ethical implications associated with its deployment.

\paragraph{Bias Propagation.}
Our framework relies on heterogeneous grounding experts (GroundingDINO, OWLv2) for region proposals. Consequently, the system may inherit or amplify biases present in these foundational detectors. If these tools exhibit performance disparities across different demographics or cultural contexts, Active-Look could inadvertently produce biased verification results.

\paragraph{Privacy and Surveillance.}
The core ``Zoom-in'' operator  is designed to resolve visual ambiguity by increasing local resolution. In real-world applications, this capability raises privacy concerns, as it could be repurposed to extract fine-grained details or Personally Identifiable Information (PII) from global scenes. Practitioners must apply strict privacy safeguards when deploying such granular inspection tools.

\bibliography{custom}

\newpage
\newpage
\appendix

\section{Experimental Settings}
\label{sec:appendix_exp_settings}

In this section, we provide a comprehensive description of the Large Vision-Language Model (LVLM) backbones used in our experiments, as well as the detailed protocols for the datasets and evaluation metrics.

\subsection{VLM Backbones}

To verify the universality of the Active-Look framework, we selected three representative LVLMs ranging from foundational baselines to state-of-the-art architectures. These models vary significantly in their visual encoders, cross-modal alignment mechanisms, and training recipes.

\paragraph{LLaVA-1.5}
LLaVA-1.5 serves as our primary baseline representing the standard projector-based architecture. It utilizes CLIP-ViT-L/14-336px as the visual encoder, which processes images at a fixed resolution of 336x336. The visual features are connected to the Vicuna-1.5 (based on Llama-2) language model through a two-layer Multi-Layer Perceptron (MLP) projector. LLaVA-1.5 is fine-tuned on a mix of academic VQA data and multimodal instruction-following data. Its fixed resolution and static encoding make it particularly susceptible to object hallucinations when visual details are small or when the reasoning chain decouples from the image embeddings.

\paragraph{Qwen3-VL}
Qwen3-VL represents the latest iteration of the Qwen-VL series, designed for high-resolution visual understanding and complex reasoning. Unlike LLaVA's fixed resolution, Qwen3-VL employs a Naive Dynamic Resolution mechanism that can process images of arbitrary aspect ratios and resolutions. It divides the input image into 14x14 patches and uses a C-Abstractor (Convolutional Abstractor) to compress visual tokens before feeding them into the Qwen3 language backbone. Qwen3-VL has been optimized for spatial reasoning and fine-grained recognition, making it a strong candidate for testing whether Active-Look can further improve performance on top of an already capable native resolution handling system.

\paragraph{InternVL2}
InternVL2 is an advanced open-source LVLM that emphasizes powerful visual representation. It employs InternViT-6B as its vision encoder, which is significantly larger than the CLIP-ViT used in LLaVA. InternVL2 utilizes a dynamic high-resolution strategy where images are tiled into 448x448 patches, allowing the model to capture intricate details. The visual tokens are aligned with the InternLM2-Chat-20B language model (or smaller variants like 8B) using an MLP projector. Despite its strong perceptual capabilities, InternVL2 still operates on a single-pass encoding paradigm during inference, meaning it cannot revisit the image to verify uncertain details once the encoding is complete, a limitation that Active-Look aims to address.

\subsection{Datasets and Evaluation Metrics}

We evaluate hallucination mitigation using three diverse benchmarks that cover binary existence verification, comprehensive multimodal perception, and free-form generation.

\paragraph{POPE}
POPE is the standard benchmark for evaluating object existence hallucination. It constructs binary Yes/No questions about objects in MSCOCO images. The benchmark consists of three splits: Random, Popular, and Adversarial. In this work, we focus on the Adversarial split, which is the most challenging. The Adversarial split queries the existence of objects that frequently co-occur with the ground-truth objects but are absent in the specific image (e.g., asking about a "keyboard" in an image containing a "mouse" and "monitor," even if the keyboard is missing). This setting strictly tests the model's ability to rely on visual evidence rather than language priors. We report Accuracy, Precision, Recall, and F1-score.

\paragraph{MME}
MME is a comprehensive benchmark comprising 14 subtasks spanning perception and cognition. To specifically assess hallucination, we utilize the four object-level perception subtasks: Existence, Count, Position, and Color. Each subtask consists of manually constructed instruction-answer pairs (mostly Yes/No questions) designed to prevent data leakage.
\begin{itemize}[itemsep=2pt,topsep=0pt,parsep=0pt]
    \item Existence: Tests whether an object exists (similar to POPE but with different data distributions).
    \item Count: Tests the model's ability to count objects correctly, which requires precise local grounding.
    \item Position: Tests spatial relationship understanding (e.g., "Is the cat to the left of the dog?").
    \item Color: Tests attribute binding capabilities.
\end{itemize}
The performance is measured using the sum of Accuracy and Accuracy+ (where Accuracy+ enforces strictly correct parsing of the answer format).

\paragraph{CHAIR}
Unlike POPE and MME which use closed-ended QA, CHAIR evaluates object hallucinations in open-ended image captioning. We generate captions for images in the MSCOCO validation set and compare the generated objects against the ground-truth annotations. The metric uses synonym mapping to match generated words to MSCOCO categories.
We report two variants:
The formulas are defined as:
\[
CHAIR_S = \frac{N_{hallucinated\_sentences}}{N_{total\_sentences}}
\]
\[
CHAIR_i = \frac{N_{hallucinated\_objects}}{N_{total\_objects\_mentioned}}
\]
Lower scores on CHAIR indicate better performance (fewer hallucinations).
\begin{table}[t]
\centering
\setlength{\abovecaptionskip}{5pt}
\setlength{\belowcaptionskip}{5pt}
\setlength{\tabcolsep}{4pt}
\renewcommand{\arraystretch}{0.5}
\begin{tabular}{llccc}
\toprule
Model & Method & $\mathcal{C}_S \downarrow$ & $\mathcal{C}_i \downarrow$ & Recall $\uparrow$ \\
\midrule
\multirow{5}{*}{LLaVA-1.5-7B}
& Baseline    & 53.0 & 18.4 & 67.0 \\
& VCD         & 51.0 & 17.5 & 70.1 \\
& OPERA        & 49.0 & 16.0 & \textbf{70.2} \\
& VAF         & 53.0 & 17.3 & 69.9 \\
& Active-Look & \textbf{15.0} & \textbf{15.8} & 64.9 \\
\bottomrule
\end{tabular}
\caption{CHAIR evaluation on MSCOCO validation set with LLaVA-1.5-7B. Lower is better for $\mathcal{C}_S$ and $\mathcal{C}_i$}
\label{tab:chair_llava_other_methods}
\end{table}

Table~\ref{tab:chair_llava_other_methods} reports CHAIR on MSCOCO with LLaVA-1.5-7B. Active-Look yields a pronounced reduction in sentence-level hallucinations, decreasing $\mathcal{C}_S$ from 53.0 to 15.0, and also lowers object-level hallucinations ($\mathcal{C}_i$: 18.4 $\rightarrow$ 15.8).
In contrast, prior decoding-time baselines (VCD, OPERA, VAF) only provide marginal improvements over the vanilla captioner, with $\mathcal{C}_S$ remaining around 49--53, indicating that hallucinated sentences are still prevalent under free-form generation.
We observe a modest Recall drop for Active-Look (67.0 $\rightarrow$ 64.9), which aligns with an evidence-priority generation pattern: by suppressing unsupported objects, the model tends to produce more conservative captions, slightly reducing coverage while substantially improving factuality.
Overall, these results suggest that conflict-triggered verification is particularly effective for reducing sentence-level hallucinations in open-ended captioning, complementing the gains observed on closed-form QA benchmarks.

Remark on comparability:
Active-Look augments the backbone captioner with conflict-triggered re-perception and evidence-priority decoding, where generation is conditioned on verified visual regions rather than purely internal priors.
Therefore, the CHAIR gains should be interpreted as the effect of introducing an explicit verification stage into open-ended captioning, instead of a pure decoding-time regularization.
We include VCD/OPERA/VAF as representative decoding-time baselines under the same backbone for reference; unlike Active-Look, they do not invoke additional region verification, which helps explain why their $\mathcal{C}_S$ reductions remain limited.

Importantly, CHAIR evaluates the end-to-end factuality of the produced captions; thus, improvements remain meaningful as they reflect the practical reliability of the overall inference pipeline.

\section{Data Collection}
\subsection{Object Scale Definitions and Data Collection}
\label{sec:appendix_scales}

As discussed in Section 4.1, the effectiveness of TwI operators is highly dependent on the granularity of the visual evidence. To rigorously analyze this trade-off, we categorize objects based on their relative area $A_{rel}$ (the ratio of the object bounding box area to the total image area) following the MSCOCO definition:

\begin{itemize}[itemsep=2pt,topsep=0pt,parsep=0pt]
    \item \textbf{Small Objects:} $A_{rel} < 10\%$. These objects typically suffer from low resolution in the original image encoder, making them prone to omission by global encoders.
    \item \textbf{Medium Objects:} $10\% \le A_{rel} \le 30\%$. Objects in this range generally balance context and detail.
    \item \textbf{Large Objects:} $A_{rel} > 30\%$. These objects usually retain sufficient semantic information in the global view. For this category, excessive zooming may be counter-productive as it can disrupt contextual relations with surrounding entities.
\end{itemize}

\paragraph{Data Collection and Balancing.}
To ensure a statistically meaningful and balanced diagnosis, we constructed a dedicated evaluation subset sampled from the GQA dataset. Unlike the long-tail distribution observed in the wild, we controlled the sample size to prevent any single scale group from dominating the metrics.
Specifically, we collected 300 distinct image--question pairs for each scale group (Small, Medium, and Large), resulting in a total of 900 diagnostic samples.
As shown in Figure~\ref{fig:gqa_scale_prompt}, to guarantee the quality of this subset, we used GPT-5 as an auxiliary assistant for sample screening and scale verification: it filters out ambiguous cases (e.g., underspecified referents) and checks whether the queried object is consistent with the bounding-box statistics that define our scale bins.
We then manually audited a random subset of GPT-5 decisions and corrected errors, ensuring that the final inclusion and scale assignment reflect the intended visual difficulty rather than artifacts of automatic filtering.
We will release the final sample IDs and scale labels to enable reproduction without relying on GPT-5.

% 在导言区添加: \usepackage{tcolorbox}
\begin{figure}[h]
    \centering
    \begin{tcolorbox}[colback=gray!5!white, colframe=gray!50!black, title=\textbf{Prompt for Object Scale Classification (GQA)}]
        \small
        \textbf{System Instruction:} \\
        You are an expert in visual analysis. Your task is to classify the size of objects in images based on their pixel area.
        
        \vspace{0.2cm}
        \hrule
        \vspace{0.2cm}
        
        \textbf{User Input:} \\
        Analyze the image and answer the following question: \textit{``\{question\}''}
        
        \vspace{0.1cm}
        \textbf{Task:} If the answer is \textbf{YES} (the object exists), classify its size based on the percentage of image area it occupies:
        
        \begin{itemize}[leftmargin=*]
            \item \textbf{Small:} Object occupies $< 10\%$ of total image area (e.g., small birds, distant objects).
            \item \textbf{Medium:} Object occupies $10\%\text{--}30\%$ of total image area (e.g., people at medium distance).
            \item \textbf{Large:} Object occupies $> 30\%$ of total image area (e.g., close-up objects, dominant subjects).
        \end{itemize}

        \vspace{0.1cm}
        \textbf{Output Format (JSON):} \\
        \texttt{\{ \\
        \hspace*{1em} "exists": true/false, \\
        \hspace*{1em} "size": "small"/"medium"/"large" (only if exists is true), \\
        \hspace*{1em} "confidence": "high"/"medium"/"low" \\
        \}}
        
        \vspace{0.1cm}
        \textit{Note: If the object does NOT exist, return \texttt{\{"exists": false\}}.}
    \end{tcolorbox}
    \caption{The prompt used to categorize GQA samples into Small, Medium, and Large scales based on visual evidence area.}
    \label{fig:gqa_scale_prompt}
\end{figure}

\begin{figure}[h]
    \centering
    \begin{tcolorbox}[colback=gray!5!white, colframe=gray!50!black, title=\textbf{Prompt for TallyQA Difficulty Classification}]
        \small
        \textbf{System Instruction:} \\
        You are an expert in visual analysis. Your task is to classify the difficulty of counting tasks in images.
        
        \vspace{0.2cm}
        \hrule
        \vspace{0.2cm}
        
        \textbf{User Input:} \\
        Analyze the image and evaluate the difficulty of this counting task: \\
        Question: \textit{``\{question\}''} \\
        Answer: \textit{``\{label\}''}
        
        \vspace{0.2cm}
        \textbf{Difficulty Classification Criteria:}
        \begin{itemize}[leftmargin=*]
            \item \textbf{Simple:} Objects are clearly visible, well-separated, with minimal occlusion (e.g., 1--3 distinct objects in clear view).
            \item \textbf{Complex:} Objects are heavily occluded, cluttered, or require distinguishing similar objects in complex scenes (e.g., dense crowds, overlapping items).
        \end{itemize}
        
        \vspace{0.2cm}
        \textbf{Response Format (JSON):} \\
        \texttt{\{ \\
        \hspace*{1em} "difficulty": "Simple"/"Complex", \\
        \hspace*{1em} "confidence": "high"/"low" \\
        \}}
        
        \vspace{0.1cm}
        \textit{Note: Analyze carefully and provide your classification.}
    \end{tcolorbox}
    \caption{The specific prompt used to categorize TallyQA samples into Simple and Complex difficulty levels based on occlusion and scene complexity.}
    \label{fig:tallyqa_prompt}
\end{figure}

\subsection{TallyQA Data Collection}
\label{sec:tallyqa_collection}

Complementary to the scale-based analysis, we extend our evaluation to the counting domain using the TallyQA dataset. Counting tasks impose a unique challenge, specifically when dealing with dense scenes. To rigorously assess the model's robustness against visual interference, we classify the samples into two difficulty levels based on the spatial arrangement and visibility of the objects.

\paragraph{Difficulty Definitions.}
Unlike standard splits that focus on question complexity, we define difficulty based on \textbf{object occlusion} and separation:

\begin{itemize}[itemsep=2pt,topsep=0pt,parsep=0pt]
    \item \textbf{Simple Samples (Isolated):} The target objects in these images are spatially separated with clear, non-overlapping boundaries. The background is relatively clean, and each instance is fully visible, making the detection task straightforward.
    \item \textbf{Complex Samples (Occluded):} This category targets scenarios with high levels of \textbf{occlusion} and dense clustering. In these samples, objects significantly overlap with one another or are partially blocked by background elements. The core challenge lies in \textbf{visual disentanglement}—the model must distinguish individual instances despite shared boundaries and partial visibility.
\end{itemize}
\begin{figure*}[h]
    \centering
    \includegraphics[width=\textwidth]{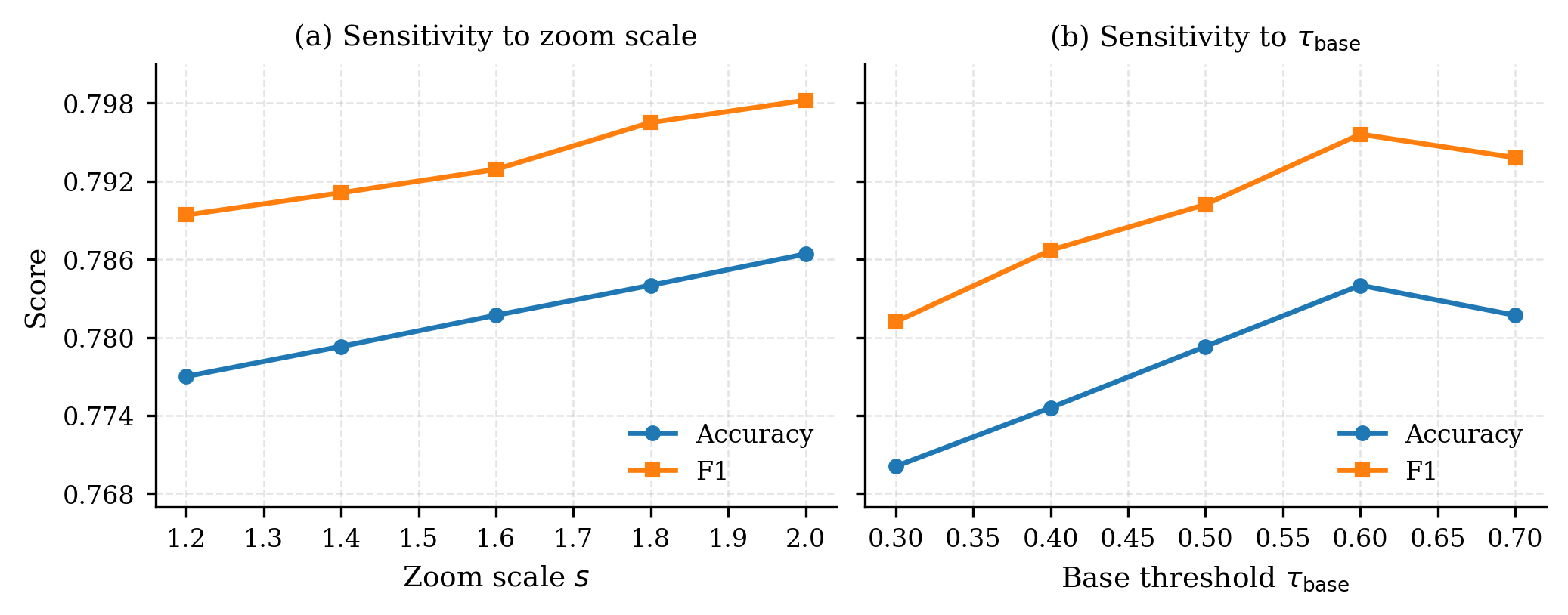}
    \caption{\textbf{Parameter sensitivity analysis.}
(a) Effect of the zoom-in scale $s$ for ROI rendering. Increasing $s$ consistently improves both Accuracy and F1, suggesting that higher effective resolution on suspicious regions benefits verification.
(b) Effect of the the base IoU threshold $\tau_{\text{base}}$ used for dual-expert consensus. Performance improves as $\tau_{\text{base}}$ increases up to $0.6$ and slightly drops at $0.7$, indicating an optimal balance between over-merging loose matches and over-splitting into disagreements.}
\label{fig:parameter_sensitivity}
\end{figure*}
As shown in Figure~\ref{fig:tallyqa_prompt}, to construct this occlusion-aware subset, we employed GPT-5 as an auxiliary assistant to refine the data selection. Since metadata regarding occlusion is not explicitly provided in the raw dataset:
\begin{enumerate}
    \item We tasked the model to analyze the image content and filter out samples where objects are clearly separated (assigning them to \textit{Simple}).
    \item We specifically selected samples where the model described the scene using occlusion-related cues such as ``crowded,'' ``stacked,'' or ``overlapping'' (assigning them to \textit{Complex}).
\end{enumerate}
We then manually audited a random subset of GPT-5 decisions and corrected errors to mitigate potential filtering bias and labeling noise.
Finally, we sampled 300 pairs for the Simple (Isolated) category and 300 pairs for the Complex (Occluded) category, ensuring a balanced evaluation of the method's performance under varying degrees of visual obstruction.

\section{Implementation Details}
\label{sec:implementation}

We implement \emph{Active-Look} using PyTorch and deploy the LVLM backbones via vLLM for efficient inference. The dual-expert system utilizes \textbf{GroundingDINO} and \textbf{OWLv2} (specifically \texttt{owlv2-base-patch16-ensemble}) as the region proposal network. The default hyperparameter configurations, determined through empirical tuning on a held-out validation set, are summarized in Table~\ref{tab:hyperparams}.

During the visual encoding stage, global images are resized to a maximum edge length of 384 pixels to balance efficiency with context retention. For the \emph{Active Stare} mechanism, we apply a zoom factor of $1.5\times$ to the region of interest (ROI) to resolve fine-grained ambiguities.

\begin{table}[h]
    \centering
    \small
    \setlength{\tabcolsep}{4pt}
    \begin{tabular}{l l c}
    \toprule
    \textbf{Category} & \textbf{Parameter} & \textbf{Value} \\
    \midrule
    \multirow{1}{*}{\textit{Visual Perception}} 
     & ROI Zoom Scale ($s$) & $1.5\times$ \\
    \midrule
    \multirow{2}{*}{\textit{Conflict Arbitration}} & Base IoU Threshold ($\tau_{\text{base}}$) & $0.6$ \\
     & Conflict Range ($\tau_{low}, \tau_{high}$) & $0.5, 0.7$ \\
    \midrule
    \multirow{2}{*}{\textit{Inference}} & LLM Temperature & $1.0$  \\
     & Seed & 42 \\
    \bottomrule
    \end{tabular}
    \caption{\textbf{Hyperparameter configurations.} These settings control the granularity of visual evidence, the sensitivity of the dual detectors, and the thresholds for the consensus-conflict arbitration mechanism.}
    \label{tab:hyperparams}
\end{table}

\section{Parameter Sensitivity Analysis}

We study the sensitivity of two key hyperparameters in Active-Look: the zoom-in scaling factor for ROI rendering and the base IoU threshold $\tau_{\text{base}}$ used in dual-expert consensus arbitration (Figure~\ref{fig:parameter_sensitivity}).

\paragraph{Zoom-in scaling factor.}
We vary the ROI scaling factor from $1.2$ to $2.0$ and report the final Yes/No performance. As the scale increases, both metrics improve steadily: Accuracy rises from $0.7770$ (scale $1.2$) to $0.7864$ (scale $2.0$), and F1 increases from $0.7894$ to $0.7982$. This monotonic trend suggests that allocating more effective resolution to the suspicious region provides consistently stronger local evidence for verification, and in our setting the potential context loss introduced by zooming does not dominate. We therefore use a default scaling factor of $1.5$ for efficiency, which achieves a strong trade-off between accuracy and computation.

\paragraph{Sensitivity to $\tau_{base}$.}
We further sweep the IoU threshold $\tau_{base}$ from $0.3$ to $0.7$. Performance improves as $\tau_{base}$ increases from $0.3$ to $0.6$ (Accuracy: $0.7701 \rightarrow 0.7840$, F1: $0.7812 \rightarrow 0.7956$), and slightly drops at $\tau_{base}=0.7$ (Accuracy $0.7817$, F1 $0.7938$). This behavior matches the role of $\tau_{base}$ in consensus partitioning: a too-small threshold may overly accept loose matches as agreement (reducing necessary verification), while an overly strict threshold can over-trigger disagreement and spend budget on marginal regions. Overall, the curve is smooth with a clear optimum around $\tau_{base}=0.6$, which we adopt as the default.

We analyze the sensitivity of two key hyperparameters in our pipeline: the zoom-in scale $s$ and the base IoU threshold $\tau_{\text{base}}$ (Figure~\ref{fig:parameter_sensitivity}).

\begin{figure}[!t]
    \centering
    \begin{tcolorbox}[colback=gray!5!white, colframe=gray!50!black, title=\textbf{Prompt for Object Extraction}]
        \small
        \textbf{System Input:} [Image] \\
        \textbf{Text Prompt:} \\
        Please analyze this image and the question: ``\textit{\{user\_query\}}''
        
        \vspace{0.2cm}
        List all objects that need to be detected to answer this question.
        
        \vspace{0.2cm}
        IMPORTANT: Output ONLY a valid JSON object in this exact format: \\
        \texttt{\{"objects": ["object1", "object2", "object3"]\}}
        
        \vspace{0.2cm}
        Do not include any explanation or additional text. Only the JSON object.
    \end{tcolorbox}
    \caption{The prompt used to extract detection targets from the user query.}
    \label{fig:prompt_extraction}
\end{figure}

\section{Prompt Templates}
\label{sec:appendix_prompts}

To ensure reproducibility, we provide the exact prompts used in our Dual-Detector Vision Agent. The system operates in three stages: 

\subsection{Stage 1: Target Object Extraction}
In the first stage (Figure~\ref{fig:prompt_extraction}), the VLM identifies which objects require visual verification based on the user's query. The prompt enforces a strict JSON output format to facilitate downstream parsing.

\begin{figure}[!t]
    \centering
    \begin{tcolorbox}[colback=gray!5!white, colframe=gray!50!black, title=\textbf{Prompt for Final Inference}]
        \small
Task: Answer the user's question with ``Yes'' or ``No'' based strictly on the visual evidence provided.

\vspace{0.15cm}
I have processed the image using advanced Object Detectors (Grounding DINO \& OWLv2).\\
Here is the visual evidence provided in IMAGE 1 (Detection Visualization):\\
\textit{\{detection\_summary\_text\}}

\vspace{0.15cm}
Image Guide:
\begin{itemize}[leftmargin=*]
    \item IMAGE 1: Detection Visualization. Green boxes = High Confidence. Red boxes = Need Further Verification.
    \item IMAGE 2: Original Image.
    \item \textit{[Conditional]} IMAGE 3: Zoomed-in view of the suspicious region for ``\textit{\{top\_suspicious\_label\}}''. This provides clearer visual evidence to verify the object.
\end{itemize}

\vspace{0.1cm}
User Question: ``\textit{\{user\_query\}}''

\vspace{0.15cm}
Decision Rules (Priority Order):
\begin{enumerate}[leftmargin=*]
    \item \textbf{Visual Evidence First}: If you can clearly see the object in the Original Image (IMAGE 2) or Zoomed Image (IMAGE 3), answer ``Yes'' regardless of the confidence label.
    \item \textbf{Green Box = Strong Signal}: If marked as ``\checkmark\ CONFIRMED'' (Green Box), it strongly suggests ``Yes''.
    \item \textbf{Red Box = Check Carefully}: If marked as ``SUSPICIOUS'' (Red Box), carefully examine IMAGE 2 and IMAGE 3. If visual evidence supports the object's presence, answer ``Yes''.
    \item \textbf{No Detection = Likely No}: If the object is not detected at all, answer ``No'' unless you can see it clearly in IMAGE 2.
\end{enumerate}

\vspace{0.1cm}
Important: The detector's confidence label (SUSPICIOUS/CONFIRMED) is a reference, not the final decision. Your answer must reflect what you actually see in the images.

\vspace{0.1cm}
Provide your answer and detailed analysis:
    \end{tcolorbox}
    \caption{The final prompt that integrates visual evidence, detection status summaries, and ROI enhancement to generate the final answer.}
    \label{fig:prompt_final}
\end{figure}

\subsection{Stage 2: Open-Vocabulary Detection}
For the extracted objects, we construct the detection prompt for Grounding DINO and OWLv2 by concatenating the object names:
\begin{quote}
    \textit{``a \{object\_1\}. a \{object\_2\}. ...''}
\end{quote}

\subsection{Stage 3: Evidence-Based Final Inference (Yes/No)}
In the final stage (Figure~\ref{fig:prompt_final}), we employ a prompt specifically tailored for Yes/No questions. The VLM receives the original image, the visual grounding results (Green/Red boxes), and optionally a zoomed-in ROI crop. The prompt explicitly instructs the model to rely on the visualized evidence to determine the final binary answer.

\begin{figure*}[!t]
    \centering
    \includegraphics[width=\linewidth]{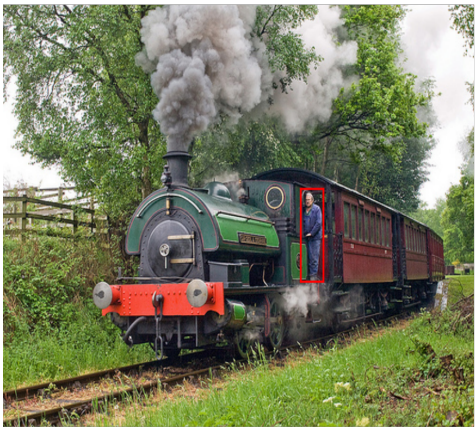}
    \caption{\textbf{Regime I: High-Confidence Validation.} 
    Query: ``Is there a baseball bat in the image?'' 
    The detector provides a strong grounding signal which aligns with the visual evidence in the global view. The model validates the high-confidence proposal without unnecessary over-correction, demonstrating efficiency in unambiguous scenarios.}
    \label{fig:case_bat}
\end{figure*}

\begin{figure*}[t]
    \centering
    \includegraphics[width=\linewidth]{Case2.pdf}
    \caption{\textbf{Regime II: Ambiguity Resolution (False Negative Correction).} 
    Query: ``Is there a person in the image?''
    The detector flags the region as \emph{Suspicious} (Red Box), which typically risks a false negative. However, the zoomed-in ROI reveals clear human features. The model prioritizes this pixel-level evidence over the low-confidence label, correctly answering ``Yes.''}
    \label{fig:case_person_yes}
\end{figure*}

\begin{figure*}[t]
    \centering
    \includegraphics[width=0.85\linewidth]{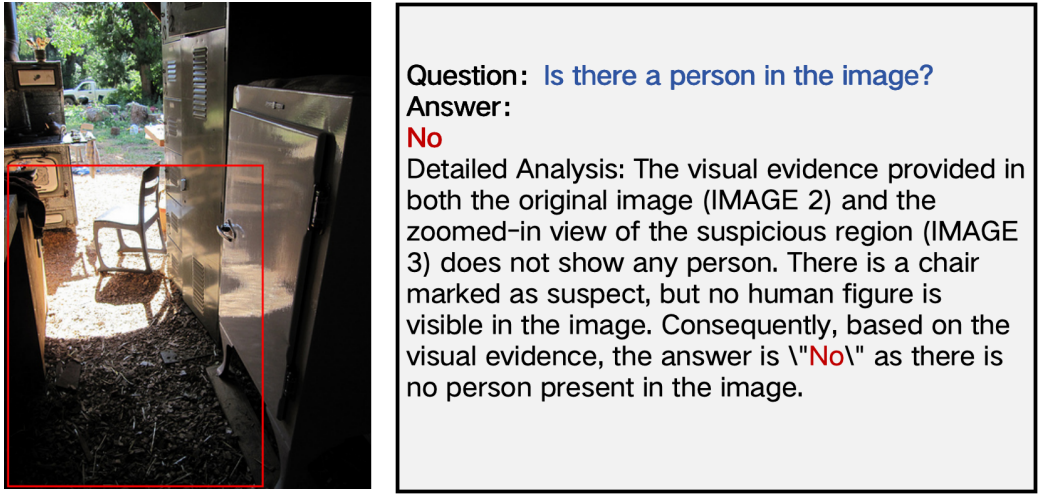}
    \caption{\textbf{Regime III: Noise Rejection (False Positive Mitigation).} 
    Query: ``Is there a person in the image?''
    The detector hallucinates a proposal on a chair-like object. Unlike standard chains that might trust the tool, our verification step reveals no human features in the ROI. The model correctly rejects the spurious proposal, answering ``No.''}
    \label{fig:case_person_no}
\end{figure*}
\section{Efficiency}
\label{sec:efficiency}
\paragraph{Token Consumption.}
Active-Look trades additional computation for improved reliability by introducing a second, evidence-seeking round when verification is triggered.
Table~\ref{tab:token_consumption} reports the token budget of LLaVA-1.5 under a 384$\times$384 setting.
Compared to a single-round baseline, the main overhead comes from extra \emph{visual input tokens} in Round~2 (1{,}152--1{,}728), corresponding to the additional auxiliary view(s) (e.g., highlighted global view and/or zoom-in ROI).
As a result, the total input increases from $\sim$726 tokens in Round~1 to $\sim$2{,}100--2{,}800 tokens across two rounds, while the text output remains small ($\sim$50--90 tokens total), indicating that the cost is dominated by visual re-encoding rather than longer generations.

We note that this overhead is \emph{budgeted} and \emph{conditional}.
Active-Look limits the number of auxiliary views and only invokes the second round when the dual experts disagree (high conflict), which concentrates computation on ambiguous cases instead of uniformly increasing cost for all samples.
In practice, this controlled token increase can be viewed as paying a small, bounded ``verification tax'' to prevent costly hallucinations in downstream use.
\paragraph{Runtime Overhead.}
Active-Look incurs additional latency due to the verification round.
On an NVIDIA H20 GPU, our end-to-end inference time is about $2.3\times$ that of standard single-pass prompting under the same evaluation setup.
This overhead is expected: once conflict is detected, the model performs an extra LVLM pass over auxiliary view(s), and the added cost is dominated by visual re-encoding, while the second-round text generation remains short.
Importantly, the slowdown is \emph{bounded} and \emph{controllable} via the view budget (e.g., number of ROIs, zoom scale, and conflict thresholds), enabling practitioners to trade efficiency for reliability.
The absolute ratio may vary with hardware, batching, and implementation optimizations; nevertheless, the dominant factor is the extra visual pass rather than longer decoding.

\begin{table*}[h]
\centering
\caption{LLaVA 1.5 Token Consumption for 384×384 Image Processing}
\label{tab:token_consumption}
\begin{tabular}{lccc}
\toprule
\textbf{Token Type} & \textbf{Round 1} & \textbf{Round 2} & \textbf{Total} \\
\midrule
\textbf{Input (Visual)} & 576 & 1,152--1,728 & \textbf{1,728--2,304} \\
\textbf{Input (Text)} & $\sim$150 & $\sim$240--350 & \textbf{$\sim$390--500} \\
\textbf{Output (Text)} & $\sim$20-40 & $\sim$30--50 & \textbf{$\sim$50--90} \\
\midrule
\textbf{Total Input} & $\sim$726 & $\sim$1,400--2,100 & \textbf{$\sim$2,100--2,800} \\
\textbf{Total Output} & $\sim$20-40 & $\sim$30--50 & \textbf{$\sim$50--90} \\
\bottomrule
\end{tabular}
\end{table*}

\section{Case Studies}
\label{sec:case_study}

We present qualitative case studies on InternVL2-8B to illustrate the decision-making dynamics of our evidence-driven inference framework. We analyze the model's behavior under three distinct regimes: (i) \emph{high-confidence validation}, where tool proposals align with obvious visual features; (ii) \emph{ambiguity resolution}, where the model must verify low-confidence proposals; and (iii) \emph{noise rejection}, where the model must discard spurious detector hallucinations.
In all instances, the model is presented with three aligned views: a detector visualization (IMAGE~1), the original global view (IMAGE~2), and a context-aware zoomed-in ROI (IMAGE~3). The core operational principle is \emph{evidence-priority decoding}: while detection confidence serves as an attentional cue, the final existence judgment is strictly grounded in pixel-level verification within the original or zoomed views.

\paragraph{Case 1: Confirmation of High-Confidence Proposals.}
Figure~\ref{fig:case_bat} depicts a scenario where the queried object (\emph{baseball bat}) is both spatially localized by the expert tools and visually salient. The grounding module provides a high-confidence confirmation, and the object requires minimal disambiguation. InternVL2-8B effectively integrates these signals, outputting ``Yes'' with a rationale that cites the correspondence between the green bounding box and the visible structure. This case demonstrates that the Active-Look framework maintains efficiency for clear samples, acting as a reliable corroborator rather than introducing unnecessary skepticism when evidence is unambiguous.

\paragraph{Case 2: Recovering Recall via Visual Verification.}
A critical failure mode in standard tool-augmented pipelines is the propagation of uncertainty. Figure~\ref{fig:case_person_yes} illustrates a case where the external detector flags a person with a \emph{suspicious} (low-confidence) label. In a naive pipeline, this might result in a conservative false negative. However, our framework leverages the zoomed-in ROI to resolve this ambiguity. Despite the detector's uncertainty, the model identifies distinct anatomical features (silhouette and limbs) in the enhanced view. By adhering to the protocol of prioritizing visual evidence over metadata labels, the model correctly recovers the object, demonstrating how active zooming can convert ambiguous proposals into decisive true positives.
\begin{table*}[!t]
\centering
\begin{tabular}{lcc}
\toprule
Partition & Sample Ratio & Baseline Error Rate \\
\midrule
Low Conflict (Consensus)     & 62\% & 12.4\% \\
High Conflict (Disagreement) & 38\% & 36.8\% \\
\bottomrule
\end{tabular}
\caption{Correlation between expert conflict and baseline error rate
on POPE Adversarial (LLaVA-1.5-7B). Samples on which the two grounding
experts disagree incur nearly $3\times$ the error rate of consensus
samples, confirming that cross-expert disagreement is a principled
proxy for visual uncertainty rather than an arbitrary selection
criterion.}
\label{tab:trigger}
\end{table*}
\paragraph{Case 3: Mitigating Hallucination via Noise Rejection.}
Conversely, Figure~\ref{fig:case_person_no} addresses the challenge of proposal noise, where the detector hallucinates a candidate region (a false positive). The region contains background clutter and a chair-like structure that mimics a human form, leading to a ``suspicious'' proposal. Crucially, the subsequent visual verification step acts as a filter. Upon inspecting the zoomed ROI, the VLM finds no semantic evidence of a human figure and overrides the proposal. This rejection mechanism highlights the robustness of the framework: rather than uncritically accepting noisy tool outputs, the model treats them merely as hypotheses to be falsified, thereby preventing object-existence hallucinations.

\paragraph{Mechanistic Interpretability and Synthesis.}
Collectively, these cases offer a mechanistic explanation for the quantitative gains reported. The decision boundary of InternVL2-8B shifts dynamically based on the clarity of visual evidence. In Case~1, the global context suffices; in Case~2, the local zoom provides the necessary granularity to boost recall; and in Case~3, the lack of visual support in the ROI prevents precision degradation. This confirms that the \emph{Active-Look} paradigm effectively decouples the region proposal step from the reasoning step, allowing the VLM to function as an independent verifier that balances the trade-off between suppressing hallucinations and maintaining coverage.
\section{Additional Experiments}

\subsection{ Verification Trigger Rate Analysis}

A natural concern with any multi-round inference framework is whether
the additional computation is broadly distributed or concentrated on a
tractable fraction of inputs. Active-Look triggers second-round
perception only when the conflict ratio $\gamma$ exceeds the adaptive
threshold $\tau_{iou}$, by design restricting overhead to samples where
heterogeneous experts genuinely disagree. Table~\ref{tab:trigger}
quantifies this behavior on POPE Adversarial (LLaVA-1.5-7B) by
reporting the baseline error rate within each partition.

The large disparity in error rates---36.8\% for high-conflict samples
versus 12.4\% for consensus samples---provides direct empirical
support for the uncertainty proxy assumed in Section~5.2. The
verification step is triggered for only 38\% of inputs, which means
that the dual-expert overhead is offset by avoiding a full second
LVLM pass on the majority of samples. This cost profile stands in
sharp contrast to exhaustive sliding-window strategies, where every
image incurs a fixed multi-view overhead regardless of perceptual
difficulty.

%-------------------------------------------------------
\subsection{ Scalability to Larger Backbones}

The diagnostic experiments in Section~4 were conducted on 7--8B parameter models, where the benefit of explicit re-perception is most apparent due to limited native visual resolution. A pertinent question
is whether the gains diminish as backbone capacity, and consequently native perceptual fidelity, increases. To investigate this, we evaluate Active-Look on InternVL2-14B and Qwen3-VL-32B under the POPE Adversarial setting; results are reported in Table~\ref{tab:large}.

\begin{table}[h]
\centering
\small
\begin{tabular}{llcc}
\toprule
Model & Method & Accuracy & F1 \\
\midrule
\multirow{2}{*}{InternVL2-14B} & Baseline    & 82.4 & 80.3 \\
                                & Active-Look & \textbf{86.2} & \textbf{83.8} \\
\midrule
\multirow{2}{*}{Qwen3-VL-32B}  & Baseline    & 85.6 & 84.9 \\
                                & Active-Look & \textbf{89.6} & \textbf{88.7} \\
\bottomrule
\end{tabular}
\caption{Active-Look on larger LVLM backbones (POPE Adversarial).
Gains are consistent across scales, indicating that conflict-driven
re-perception addresses residual visual ambiguities that stronger
native encoders do not eliminate through single-pass encoding alone.}
\label{tab:large}
\end{table}

Consistent improvements across both models argue against the hypothesis that Active-Look becomes redundant once backbone capacity is sufficient.
Qwen3-VL-32B, equipped with a high-resolution dynamic encoding
mechanism, already achieves a competitive 85.6\% baseline, yet
Active-Look still yields a 4.0-point gain. We attribute this to a
structural limitation shared by all single-pass architectures: once the
image is encoded, subsequent reasoning proceeds entirely in language
space with no opportunity to revisit uncertain visual regions. Active
re-perception directly addresses this bottleneck and, as the results
suggest, does so in a manner that is orthogonal to the improvements
afforded by scaling the backbone itself.

\section{Future Work}
\label{sec:future_work}

While \emph{Active-Look} establishes a robust baseline for inference-time hallucination mitigation, several avenues remain for extending the paradigm of conflict-driven visual verification.

\paragraph{End-to-End Active Perception.}
Currently, our framework operates in a training-free manner, relying on external, frozen grounding experts to propose candidate regions. This dependency imposes an upper bound on performance defined by the recall of these off-the-shelf tools. Future research should explore end-to-end training strategies where the LVLM learns to generate its own ``visual queries'' or control the zooming mechanism directly via differentiable attention. By internalizing the proposal step, the model could align visual acquisition more tightly with its internal reasoning state, potentially capturing abstract or long-tail concepts that generic object detectors miss.

\paragraph{Efficiency-Aware Dynamic Computation.}
The parallel execution of dual experts and the encoding of auxiliary zoomed views introduce inference latency, presenting a trade-off between faithfulness and speed. Future work will investigate \emph{adaptive computation} techniques, such as early-exit mechanisms or lightweight ``scout'' models, to determine when expensive verification is strictly necessary. Furthermore, optimizing the token usage of high-resolution crops---perhaps through adaptive token merging or compressive sensing---could reduce the computational overhead of the verification phase without compromising the granularity required for fine-grained perception.

\paragraph{Beyond Object Existence: Spatial and Temporal Reasoning.}
Our current arbitration mechanism is optimized primarily for object existence and attribute verification. However, hallucinations in LVLMs often manifest as erroneous spatial relationships (e.g., ``left of'' vs.\ ``right of'') or misidentified actions. Extending \emph{Active-Look} to these domains requires new forms of visual evidence beyond static crops, such as relational highlighting or temporal segment retrieval in video inputs. We aim to generalize the conflict-detection logic to these complex reasoning tasks, enabling the system to actively verify structural and temporal hypotheses.
\end{document}